\documentclass[runningheads]{llncs}
\usepackage[T1]{fontenc}
\usepackage{graphicx}
\usepackage[ruled, lined, longend, linesnumbered]{algorithm2e}
\usepackage{float}
\usepackage{amssymb}
\usepackage{booktabs}
\usepackage{colortbl}
\usepackage{xcolor}
\usepackage{orcidlink}
\usepackage[misc,geometry]{ifsym}

\begin{document}

\title{XC-NAS: A New Cellular Encoding Approach for Neural Architecture Search of Multi-Path Convolutional Neural Networks}

\titlerunning{XC-NAS: New Cellular Encoding NAS}

\author{Trevor Londt\inst{1}$^{\textrm(\textrm\Letter\textrm)}$\orcidlink{0000-0002-9428-8089} \and Xiaoying Gao\inst{1}\orcidlink{0000-0002-6326-7947} \and Peter Andreae\inst{1}\orcidlink{0000-0002-2789-680X}
\and Yi Mei\inst{1}\orcidlink{0000-0003-0682-1363}}

\authorrunning{T. Londt et al.}

\institute{Centre for Data Science and Artificial Intelligence \& School of Engineering and Computer Science, Victoria University of Wellington,\\ PO Box 600, Wellington 6140, New Zealand \\
\email{\{trevor.londt,xiaoying.gao,peter.andreae,yi.mei\}@ecs.vuw.ac.nz}}

\maketitle  

\begin{abstract}
Convolutional Neural Networks (CNNs) continue to achieve great success in classification tasks as innovative techniques and complex multi-path architecture topologies are introduced. Neural Architecture Search (NAS) aims to automate the design of these complex architectures, reducing the need for costly manual design work by human experts. Cellular Encoding (CE) is an evolutionary computation technique which excels in constructing novel multi-path topologies of varying complexity and has recently been applied with NAS to evolve CNN architectures for various classification tasks. However, existing CE approaches have severe limitations. They are restricted to only one domain, only partially implement the theme of CE, or only focus on the micro-architecture search space. This paper introduces a new CE representation and algorithm capable of evolving novel multi-path CNN architectures of varying depth, width, and complexity for image and text classification tasks. The algorithm explicitly focuses on the macro-architecture search space. Furthermore, by using a surrogate model approach, we show that the algorithm can evolve a performant CNN architecture in less than one GPU day, thereby allowing a sufficient number of experiment runs to be conducted to achieve scientific robustness. Experiment results show that the approach is highly competitive, defeating several state-of-the-art methods, and is generalisable to both the image and text domains.

\end{abstract}

\section{Introduction}
Designing state-of-the-art CNNs for a classification task is not trivial; it requires expert skills and costly development time. Furthermore, training times have progressively increased as CNN architectures have become more complex. Researchers have focused on developing Neural Architecture Search (NAS) algorithms that automatically design CNNs without human intervention to mitigate these problems. One such variant approach, Evolutionary Computation-based Neural Architecture Search (ECNAS) \cite{Zhan2022EvolutionarySurvey}, has shown success in automatically designing CNN architectures for various classification tasks. However, many ECNAS algorithms are typically restricted to only one problem domain and task, such as image or text. Due to the high computational cost, time, and intractable size of the search space involved \cite{Zhou2021ASearch}, ECNAS algorithms typically evolve the micro-architecture and treat the macro-architecture exclusively as layers of micro-architectures stacked together in series to form a complete CNN architecture. The theory is that training a micro-architecture representing a simple computational block takes less time than training an entire CNN macro-architecture at once. The assumption is that this evolved micro-architecture can be stacked to produce a single-path CNN macro-architecture. Such works have succeeded in evolving highly performant CNNs, however, this approach prevents the possibility of exploring multi-path CNN macro-architecture topologies. Multi-path CNN macro-architectures can allow multiple feature reuse between various layers of the network or provide different processing paths through the network representing network ensembles, which may boost classification performance. Inspired by the capabilities of Cellular Encoding (CE) \cite{Gruau1994} to evolve multi-pathed topologies, this work proposes a new, configurable, and extensible CE representation (XC) with the ability to represent novel multi-pathed CNN architectures. Furthermore, the proposed encoding incorporates the ability to dynamically modify the depth and width of a CNN architecture during the evolutionary process to allow the creation of architectures of varying widths and depths. Since CE is involved with representing a sequence of operations to be executed that are encapsulated in a tree structure, Genetic Programming (GP) \cite{Poli2008AProgramming} is used to facilitate the evolution of CNN architectures using the proposed CE representation because GP itself is designed for running genotypes as programs in a tree data structure. The proposed CE representation forms the backbone of evolving multi-pathed macro-architectures for the proposed algorithm in this work. Experiments on commonly used image and text classification benchmark datasets show competitive results compared to the state-of-the-art models across all datasets. The contributions of this work are:
\begin{enumerate}
    \item A new configurable CE representation (XC), including supporting network structures, that is designed to represent multi-pathed CNN architectures for image or text classification tasks, with the ability to dynamically adjust the width and depth of the architecture during its construction.
    \item Three new ECNAS algorithms using the XC representation to evolve multi-path CNN architectures, each making use of a different handcrafted micro-architecture, namely a simple convolutional block, a ResNet block, or an Inception module. The proposed algorithms can evolve multi-path CNN architectures for image or text classification tasks in less than one GPU day.
\end{enumerate}

\section{Background}
Cellular Encoding \cite{Gruau1994}, inspired by the division of biological cells to produce complex organisms, is a generative encoding strategy to synthesise artificial neural network topologies. CE genotypes, represented as tree structures, are encoded as a sequence of instructions applied on an initial \emph{ancestor network} containing a singular \emph{cell}. Each execution of an instruction transforms the ancestor network into a different topology by adding, modifying, or removing cells in the ancestor network. This approach allows the neural network to grow organically until a suitably sized network topology with enough capacity for the problem under consideration has been generated. The instruction set of cellular encoding is extensive, however, the essential instructions are related to the duplication of a cell in the network. Sequential (SEQ), parallel (PAR), and recursion (REC) are the main duplication instructions. Sequential duplication indicates that a cell being operated on will divide into two cells, each connected in series. Parallel duplication is the division of a cell into two cells that are connected in parallel. REC operations are only executed once and repeat a sequence of instructions in the genotype. Terminal operations are the END operation, a leaf node in the genotype. The original CE encoding was used in a Genetic Algorithm \cite{Holland1992GeneticAlgorithms} (GA) to facilitate evolutionary operations. However, modern variants of CE \cite{Londt2021EvolvingProgramming,Broni-Bediako2020EvolutionaryEncoding} are used in Genetic Programming \cite{Koza1994GeneticSelection} (GP) algorithms because GP is a population-based EC method that operates directly on tree-based genotypes representing executable programs, making it a perfect and convenient fit for CE-based genotypes.

EC-based NAS (ECNAS) \cite{Zhan2022EvolutionarySurvey} is an evolutionary-based approach for searching performant network architectures. ECNAS uses an evolutionary algorithm with appropriate representation to explore the space of possible network architectures. The representation aims to constrain the search space size. The training of each potential network architecture under consideration typically uses backpropagation. The fitness of a trained network influences its survivability for the next generation in the evolutionary process. ECNAS algorithms have succeeded in evolving network architectures for many domains, including complex tasks such as image and text classification. Various EC-based search methods have been used in proposed ECNAS algorithms, including GA \cite{Holland1992GeneticAlgorithms}, Particle Swarm Optimisation (PSO) \cite{KennedyParticleOptimization}, and CE \cite{Gruau1994}.

The approaches taken by GA- \cite{Xie2017GeneticCNN} and PSO-based \cite{FernandesJunior2019ParticleClassification} algorithms typically search for CNNs containing single-path macro-architectures, missing the opportunity to explore novel multi-path architectures - which may offer added benefits of feature reuse and ensemble qualities to boost performance further. CE-based approaches can potentially explore unknown multi-path architecture search spaces. Londt et al. \cite{Londt2021EvolvingProgramming} have successfully used a CE-based approach to evolve multi-path CNN architecture for text classification tasks achieving competitive state-of-the-art results. However, their algorithm has not been designed or implemented for image classification tasks and does not consider modifying the width or depth of the network during evolution. Broni-Bediako et al.'s \cite{Broni-Bediako2020EvolutionaryEncoding} work focused on a CE-based approach to perform evolution in the local space, evolving micro-architectures that are stacked to create a single path macro-architecture, achieving impressive performance results against peer-competitors, again showing the effectiveness and utility of CE. However, their approach does not consider the global search space to evolve multi-path macro-architectures for CNNs, leaving a gap in the literature that needs to be investigated. To address these limitations, a new CE encoding (XC) is proposed and implemented in a new ECNAS algorithm, XC-NAS, capable of evolving multi-path CNN architectures for image or text classification tasks of varying widths and depths.
 
\section{XC-NAS: Cellular Encoding for NAS}

\subsection{Framework}
\begin{figure}[H]
    \centering
    \resizebox{\linewidth}{!}
    {
    \includegraphics{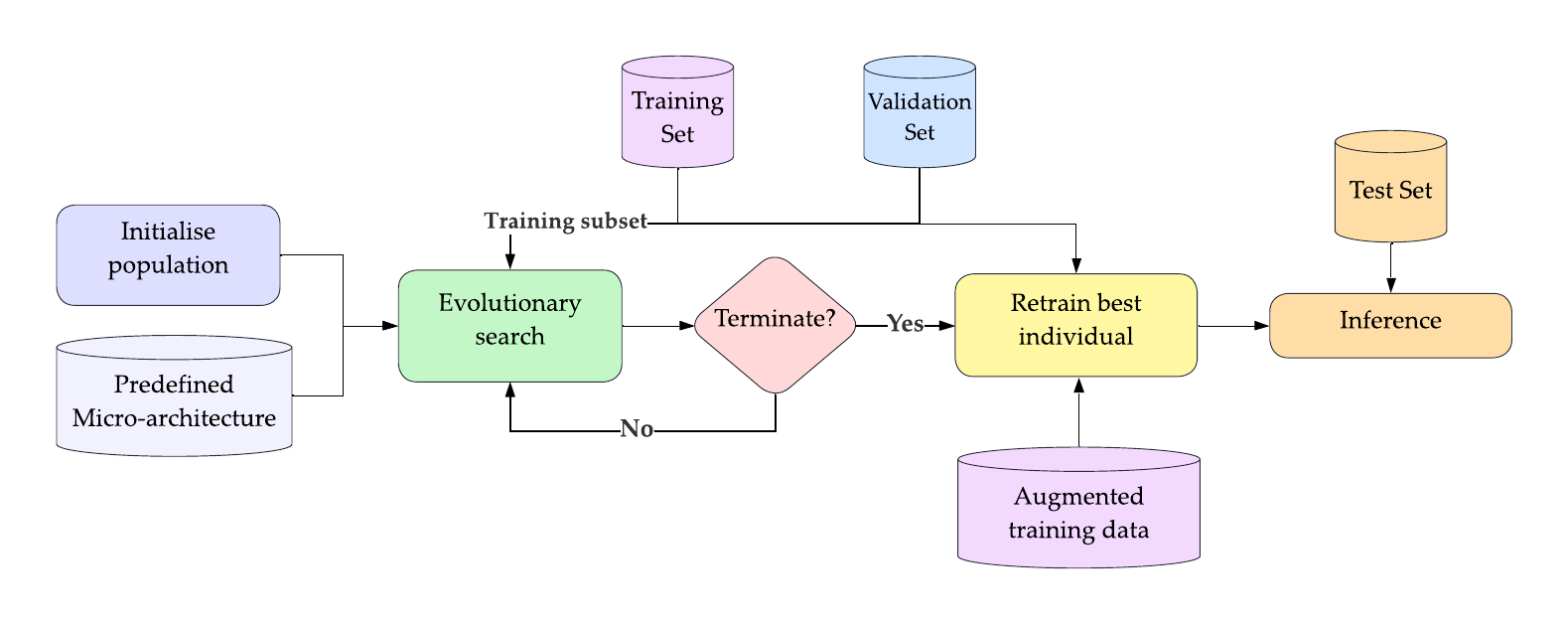}
    }
    \caption{Framework of XC-NAS.}
    \label{fig:xcnas}
\end{figure}

The general process of the proposed algorithm is presented in Fig. \ref{fig:xcnas}, and a detailed listing is provided in Algorithm \ref{alg:algo}. A subset of the training set is generated based on a predetermined percentage. A surrogate model approach is taken where only a subset of the training dataset is used during the evolutionary process as is done in \cite{Broni-Bediako2020EvolutionaryEncoding,Londt2021EvolvingProgramming}, thereby evolving low-resolution models, after which the best-evolved model is retrained using the entire training set to produce a higher resolution model for conducting inference. This approach is taken to decrease the evolutionary training time significantly. 
Each individual in the population is generated based on a randomly selected depth between a predefined minimum and maximum value.
\vspace{-0.5cm}
\vspace{-0.9cm}
\noindent Each individual in the population is decoded to a phenotype (CNN) and uploaded to the graphics processing unit (GPU). 
If the phenotype is too large to fit into the memory of the GPU, then the phenotype is considered unfit for the environment and assigned a fitness value of -1.0. This approach ensures that the phenotype will not survive into the next generation. The phenotype is trained using the generated training set and validation set, with the resulting fitness being returned when the training process terminates. The evolutionary operations of elitism, selection, crossover, and mutation are then conducted. A new population is generated, and the evolutionary process repeats until the required generations are reached. Finally, the fittest genotype in the population is decoded into a phenotype and trained using the entire training set, validation set, and data augmentation.
\begin{center}
\begin{algorithm}[H]
    \label{alg:algo}
    \caption{Pseudocode of XC-NAS.}
    \textbf{begin}\;
    \quad$seed \leftarrow random()$\;
    \quad$\mathcal{D}_{train}\leftarrow subset(seed,\mathcal{D}_{full\_train})$\;
    \quad$\mathcal{P} \leftarrow init(min, max)$\;
    \While{$not\ maximum\ generations$}{%
        \ForEach{$\kappa \in \mathcal{P}$}
        {
            $\rho \leftarrow decode(\kappa, GPU)$\;
            \eIf{$\rho \notin GPU$}
            {
                $val_{acc} \leftarrow -1.0$\;
            }
            {
                $val_{acc} \leftarrow evaluate(\rho,\mathcal{D}_{train},\ \mathcal{D}_{val})$\;
            }
        }
        $\varepsilon \leftarrow elite \subset \mathcal{P}$\;
        $\varsigma \leftarrow tournament(\mathcal{P})$\;
        $\phi \leftarrow crossover(\varsigma )$\;
        $\lambda \leftarrow mutate(\phi)$\;
        $ \mathcal{P} \leftarrow limit(\lambda \cup \varepsilon)$;
     }
     $\beta \leftarrow fittest \in \mathcal{P}$\;
     $\rho \leftarrow decode(\beta, GPU)$\;
     $test_{acc} \leftarrow evaluate(\rho,\mathcal{D}_{train},\ \mathcal{D}_{val}, \mathcal{D}_{test}, \mathcal{D}_{augment})$\;
    \textbf{end}\;
\end{algorithm}
\end{center}

\subsection{Ancestor Network, Cell, and Micro-architecture}

\begin{figure}[H]
    \centering
    \scalebox{0.6}
    {
    \includegraphics{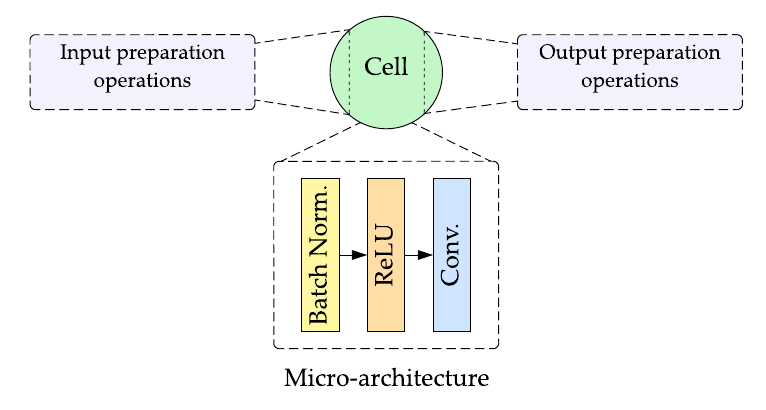}
    }
    \caption{Proposed cell structure.}
    \label{fig:cell}
\end{figure}

The network cell is the fundamental unit in a phenotype produced by a CE-based algorithm. We adopt a similar approach proposed by Londt et al. \cite{Londt2021EvolvingProgramming}, where a cell encapsulates a predefined micro-architecture. Fig. \ref{fig:cell} presents the proposed cell configuration. The cell configuration comprises three components: an input section, a \emph{block} section containing a micro-architecture, and an output section. The input section includes operations that transform the incoming connections and data from preceding cells as required by CE operations, for example, increasing or decreasing filter counts. Similarly, the output section of the cell also allows specified operations to be executed during run-time to transform the cell's outgoing connections and output data as required. The middle or block section contains the micro-architecture of the cell.

In this paper, three different micro-architectures are used and evaluated, namely Simple, ResNet-like, and InceptionNet-like micro-architectures. The Simple micro-architecture serves as a baseline micro-architecture, containing a batch normalisation layer (BN) followed by a ReLU activation layer and a convolutional layer.
ResNet \cite{He2016DeepRecognition} and InceptionNet \cite{Szegedy2015GoingConvolutions} are highly performant CNN architectures, each making a significant contribution to the literature, and are therefore leveraged in this work. The ResNet-like micro-architecture is the same as the pre-activation block in \cite{He2016DeepRecognition}. The Inception-like micro-architecture replicates the Inception module proposed by Szegedy et al. \cite{Szegedy2015GoingConvolutions}. The convolution type will be temporal for text classification tasks or spatial for image classification tasks. 

All genotypes in the population have a corresponding phenotype representing a CNN architecture. The initial phenotypes of each genotype are based on an \emph{ancestor network}. We define the ancestor network in Equation \ref{eq:an}.

\begin{equation}\label{eq:an}
\Phi(\varsigma,\varepsilon,\psi): \psi(\varepsilon(\varsigma(\beta))) \rightarrow \mathbb{Z}_{1}^n
\end{equation}

\noindent $\varsigma$ is the stem, $\varepsilon$ is the feature extractor, $\psi$ is the classifier, and $\beta$ is a batch of input data which is mapped to a class $i\in\mathbb{Z}_{1}^n$.

The feature extractor, $\varepsilon$, is defined in Equation \ref{eq:fe}
\begin{equation}\label{eq:fe}
\tau(C, E) \rightarrow \varepsilon
\end{equation}

\noindent $C$ is the set of cells in the network, $E$ is a set of edges between the cells, and $\tau$ is a mapping of the edges and cells to produce a topology representing the feature extractor.
The initial feature extractor contains only one cell. The evolutionary process primarily involves the feature extractor part of the ancestor network, creating more cells and interconnections between them to form the macro-architecture. The architecture of the stem in the ancestor network depends on the domain in which the ancestor network will operate.
\subsection{Encoding Scheme}
\begin{figure}[H]
    \centering
    \resizebox{\linewidth}{!}{
        \includegraphics{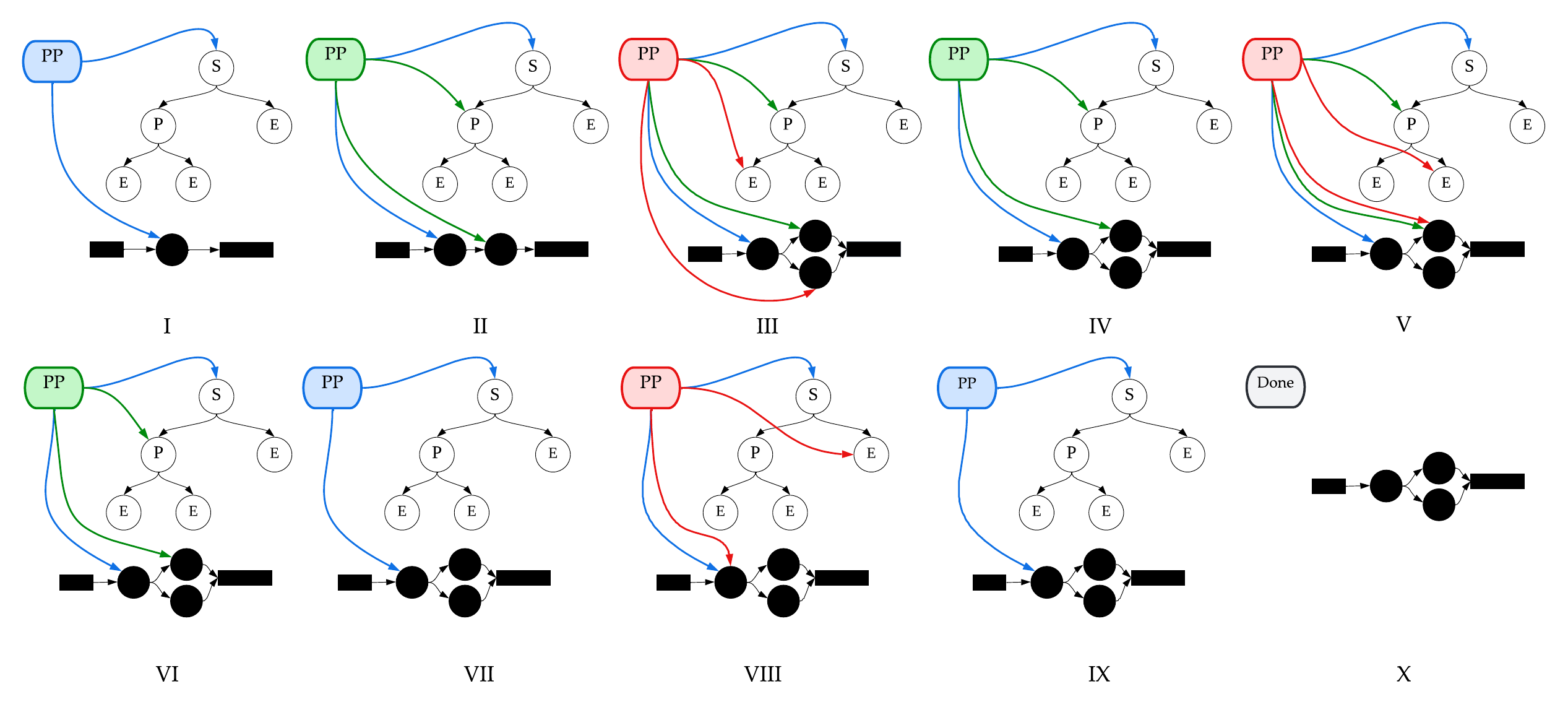}
    }
    \caption{Growing a neural network from an ancestor network. \textbf{PP}=Program Pointers, \textbf{S}=Sequential Split, \textbf{P}=Parallel Split, \textbf{E}=End Terminal.}
    \label{fig:depth_first_cellular_encoding}
\end{figure}

The proposed algorithm uses a depth-first traversal approach. Fig. \ref{fig:depth_first_cellular_encoding} presents an example of the program execution of a genotype, represented by a program tree structure, operating directly on an ancestor network to construct a final CNN architecture. Two key CE operations are demonstrated namely the sequential (\textbf{S}) and parallel (\textbf{P}) split operations.
\vspace{-0.1cm}
\vspace{-0.25cm}
\noindent In step \textbf{I}, the program pointer (PP), displayed in blue, points to the first program symbol, an \textbf{S} symbol, and the cell that will be operated on in the ancestor network.
Step \textbf{II} represents the program's state after executing the \textbf{S} symbol. The cell in the ancestor network is divided into two cells connected in series. The PP retains the original pointers, in blue, to the \textbf{S} symbol and its corresponding cell it operated on. These pointers are maintained to facilitate the back tracing of the tree structure. A new green pointer points to the next program symbol, \textbf{P}, including the newly added cell on which the P symbol will operate.
Step \textbf{III} displays the results of the execution of the \textbf{P} symbol. The cell has divided, and the resulting two cells are connected in parallel. The remainder of the program, from steps IV to IX, represents the results of executing the E symbols, which do not affect the ancestor neural network. 
In step \textbf{X}, the final architecture of the CNN is presented.

Two critically important hyperparameters of a CNN are its depth and width. The depth refers to the longest path of consecutive convolutional layers in the network from the input to the output. The width refers to the number of filters at a specific layer in the network. A new Cellular Encoding representation (XC) is proposed to encapsulate these two critical hyperparameters to allow the evolution of CNNs of varying depth and width. A cell that is being operated on is referred to as a mother cell. Any cell produced from a mother cell is called a child cell. An explanation of the proposed cellular encoding operations is provided below:

\begin{enumerate}
    \item \textbf{S}: the child cell inherits the mother cell's number of filters and depth. The child cell is inserted into the network after the mother cell.
    \item \textbf{SID}: the child cell inherits the mother cell's number of filters but increases the depth of the child cell by one block compared to the mother cell's depth.
    \item \textbf{SDD}: the child cell inherits the mother cell's number of filters but decreases the depth of the child cell by one block compared to the mother cell's depth. If the mother cell's depth is one block, then there will be no decrease, and the child cell will have a depth of one block.
    \item \textbf{SIW}: the child cell inherits double the number of filters of the mother cell and the same depth value as the mother cell.
    \item \textbf{SDW}: the child cell inherits half the number of filters of the mother cell and the same depth value. A minimum filter count is predefined; if it is breached, the child cell will inherit the minimum filter count value.
    \item \textbf{S*D*W}: the sequential split operation will inherit the mother cell's attributes and either increase or decrease the width and depth accordingly, as discussed in the previous operations. For example, SIDDW would perform a sequential split, increase the child cell's depth and also decrease the width of the child cell.
    \item \textbf{P*}: the equivalent P operations behave the same as the S operations except that the mother and child cells are connected in parallel, and the outputs from both cells are concatenated channel-wise. For example, PID would connect the mother and child cell in parallel and increase the child cell's depth.
\end{enumerate}
Table \ref{tab:extended_cellular_encoding} lists the new encoding operations and associated configurable hyperparameters. Each operation has two hyperparameters that can be adjusted to alter the magnitude of the effect an operation has on a cell. The first hyperparameter is the \emph{depth adder}. The value "1" adds an extra block inside the cell. The value "-1" will remove a block from the cell unless only one block is left, in which case nothing will be done. The second hyperparameter is the \emph{width multiplier}. A value of "2" indicates a doubling of the number of inherited filters, and a value of "0.5" will halve the number of inherited filters.
\begin{table}[H]
    \caption{\textbf{ID}=Increase Depth, \textbf{DD}=Decrease Depth, \textbf{IW}=Increase Width, \textbf{DW}=Decrease Width.}
    \centering
        \begin{tabular}{|c|c|c|c|c|c|}
            \hline
            Operation  & Depth + & Operation & Width $\times$\\
            \hline
            ID & 1 & IW & 2 \\
            \hline            
            DD & -1 & DW & 0.5 \\
            \hline
        \end{tabular}
    \label{tab:extended_cellular_encoding}
\end{table}

\subsection{Architecture Evolution}
A uniform single-point mutation operation is chosen for the proposed algorithms. The operation involves randomly selecting a node in the genotype and replacing the sub-tree at that node with a randomly generated subtree of a specified size. The single-point crossover operation is chosen for its simplicity of operation and implementation. This operation works by randomly selecting a point in each parent genotype and crossing over the sub-tress located at these points in each parent to form two offspring genotypes. Tournament Selection is an appropriate initial choice for the proposed algorithms and provides a reasonable balance between exploration and exploitation, which the tournament size can control. After training a candidate phenotype using stochastic gradient descent (SGD), the phenotype's best-recorded validation accuracy is used to determine its corresponding genotype's fitness. Early stopping is implemented to perform regularisation and prevent wasting compute time. The validation accuracy at the end of each training epoch is monitored, and if the accuracy has not improved beyond a predefined delta for a specified number of epochs, training is stopped.

\section{Experiment Design}
\subsection{Datasets and Peer Competitors}
To determine the proposed approach's effectiveness and generalisability, we consider classification tasks from two disparate domains, namely the image and text domains. Five commonly used datasets are selected. Two datasets are from the text domain, and three are from the image domain. The AG's News \cite{10.5555/2969239.2969312} dataset is considered a medium-difficulty dataset, whereas the Yelp Reviews Full \cite{Conneau2017VeryClassification} dataset is considered high-difficulty. KMNIST \cite{Clanuwat2018DeepLiterature} is an image dataset containing handwritten characters and bridges the text and image domains in this work. The Fashion-MNIST \cite{Xiao2017Fashion-MNIST:Algorithms} dataset is chosen as a more challenging dataset than KMNIST. Finally, CIFAR-10 \cite{Krizhevsky2009LearningImages} is a popular benchmark dataset which is considered to be of medium difficulty to model.
\noindent Several state-of-the-art peer competitors are chosen for benchmarking, including both manually designed architectures and NAS algorithms. For the text domains, Zhang et al.'s \cite{10.5555/2969239.2969312} original manually designed character-level model, Conneau et al.'s \cite{Conneau2017VeryClassification} manually designed VDCNN model, and the GP-Dense \cite{Londt2021EvolvingProgramming} NAS algorithm are compared against. For the image domain, ResNet-110 \cite{He2016DeepRecognition} and InceptionNet \cite{Szegedy2015GoingConvolutions} architectures are chosen as the expert-designed manual models and numerous NAS algorithms, including the CE-based CE-GeneExpr \cite{Broni-Bediako2020EvolutionaryEncoding}, FPSO \cite{Huang2021ADesign}, EvoCNN \cite{Sun2020EvolvingClassification}, CGP-CNN \cite{Suganuma2017AArchitectures}, and EIGEN \cite{Ren2019Eigen:Scratch} are chosen for benchmarking.
\subsection{Parameter Settings}
The parameter settings used for the experiment are based on those used in \cite{Broni-Bediako2020EvolutionaryEncoding,Londt2021EvolvingProgramming}. The number of generations and the population size are set to 20. Genotypes are randomly initialised to depths between 2 and 17, inclusively.
\begin{table}[H]
  \caption{Parameter settings for CNN training process.}
  \centering
\resizebox{0.85\linewidth}{!}{
  \begin{tabular}{l{r}l{r}}
  \toprule
    Parameter & Setting & Parameter & Setting \\
    \midrule
    Number of generations & 20 & Population size & 20\\
    Genotype depth & [2,17] & Elitism & 0.1 \\
    Crossover probability & 0.5 & Mutation probability & 0.01 \\
    Mutation tree growth size & [1,3] & Tournament size & 3 \\    
    Epochs & Evol=15, Full=300 & Batch size & 128 \\
    Initial learning rate & 0.01 & Momentum & 0.9 \\
    Learning schedule & Evol.=\cite{Londt2021EvolvingProgramming}, Full=\cite{He2016DeepRecognition}& Weight Init. & Kaiming.\cite{He2016DeepRecognition} \\
    Training data usage & 0.25 & Experiment runs & 30 \\
    \bottomrule
  \end{tabular}
}
  \label{tab:training_parameters}
\end{table}
\noindent The evolutionary and network training settings are those commonly used in the community. The training set used during the evolutionary process is set at 25\% of the entire training set. Three different micro-architectures are tested with XC-NAS, and thirty independent experiment runs are conducted for each configuration.

\section{Experimental Results and Discussions}
The best test results recorded for each of the 30 independent runs, and the peer competitors, are listed in Table \ref{tab:test_Results}. On the text datasets, it can be seen that XC-NAS (Inception) has outperformed the current state-of-the-art CE-based competitor, GP-Dense. XC-NAS (Inception) has only slightly underperformed VDCNN-Convolution on the AG's News dataset. On the Yelp Reviews Full dataset, XC-NAS (ResNet) appears to have performed the best of the XC-NAS variants. Regardless, XC-NAS (ResNet) has outperformed the CE-based GP-Dense competitor but underperformed the manually designed architectures, demonstrating that the Yelp Reviews Full dataset is challenging to model automatically.
\begin{table}[H]
\caption{Test accuracies (\%) compared to peer competitors. (M = Manual, A = NAS, n/a = Not Applicable, '-' = Not Available, best = Blue, second best = Orange)}
\centering
  \resizebox{\linewidth}{!}{
    \begin{tabular}{lrrrrrr}
        \toprule
        \multicolumn{1}{l}{Algorithm} & \multicolumn{1}{l}{Type} & \multicolumn{1}{l}{AG's News}& \multicolumn{1}{l}{Yelp Reviews Full} & \multicolumn{1}{l}{KMNIST} & \multicolumn{1}{l}{Fashion-MNIST} & \multicolumn{1}{l}{CIFAR-10} \\
        \toprule
        XC-NAS (Simple) & A & 90.71 & 60.92 & 98.42 & 94.31 & 93.24 \\
        XC-NAS (ResNet) & A & 89.94 & 61.26 & \textcolor{orange}{\textbf{98.68}} & 94.72 & 93.74 \\
        XC-NAS (Inception) & A & \textcolor{orange}{\textbf{91.15}} & 61.01 & \textcolor{blue}{\textbf{99.13}} & \textcolor{blue}{\textbf{95.39}} & \textcolor{orange}{\textbf{94.85}} \\
        \arrayrulecolor{black!30}\midrule        
        Zhang et al. Lg. Full Conv.\cite{10.5555/2969239.2969312} & M & 90.15 &  \textcolor{orange}{\textbf{61.60}} & n/a & n/a  & n/a \\
        VDCNN-Convolution \cite{Conneau2017VeryClassification} & M & \textcolor{blue}{\textbf{91.27}} & \textcolor{blue}{\textbf{64.72}} & n/a & n/a  & n/a \\ 
        GP-Dense \cite{Londt2021EvolvingProgramming} & A & 89.58 & 61.05 & n/a & n/a  & n/a \\ 
        \arrayrulecolor{black!30}\midrule        
        ResNet-110. \cite{He2016DeepRecognition} & M & n/a & n/a  & 97.82 & 93.40 &  93.57 \\        
        InceptionNet (GoogLeNet) \cite{Szegedy2015GoingConvolutions} & M & n/a & n/a &  97.95 & 92.74 & 93.64\\    
        EvoCNN \cite{Sun2020EvolvingClassification} & A & n/a  & n/a & - & 94.53 & - \\                
        FPSO \cite{Huang2021ADesign} & A & n/a  & n/a & - & \textcolor{orange}{\textbf{95.07}} & 93.72 \\
        CGP-CNN-ConvSet \cite{Suganuma2017AArchitectures} & A & n/a  & n/a & - & - & 93.25 \\
        EIGEN \cite{Ren2019Eigen:Scratch} & A & n/a  & n/a & - & - & 94.60 \\  
        CE-GeneExpr \cite{Broni-Bediako2020EvolutionaryEncoding} & A & n/a  & n/a & - & - & \textcolor{blue}{\textbf{96.26}}\\
        \hline
    \end{tabular}
    }
    \label{tab:test_Results}
\end{table}
\vspace{-0.5cm}
On the KMNIST and Fashion-MNIST datasets, XC-NAS (Inception) has outperformed all peer competitors. On CIFAR-10, XC-NAS (Inception) has surpassed all peer competitors except for the CE-GeneExpr algorithm. The results show the XC-NAS (Inception) is highly competitive and generalises well across both the image and text domains.

The number of GPU days required to evolve the best model is presented in Table \ref{tab:gpu_days}. Only the peer competitors with available GPU day information available have been shown. XC-NAS has taken less than one GPU day to find the best architecture for any dataset. It is clear that the approach of XC-NAS is efficient in terms of computational costs.
\begin{table}[H]
\caption{Number of GPU days required to run the algorithm and evolve the best model compared to peer competitors.}
\centering
  \resizebox{0.5\linewidth}{!}{
    \begin{tabular}{lr}
        \toprule
        Algorithm  & GPU Days \\
        \midrule
        XC-NAS (Each dataset) & $< 1$ \\
        \arrayrulecolor{black!30}\midrule
        EvoCNN \cite{Sun2020EvolvingClassification} (Fashion-MNIST) & 4\\
        CGP-CNN-ConvSet \cite{Suganuma2017AArchitectures} (CIFAR-10) & 29.8 \\
        EIGEN \cite{Ren2019Eigen:Scratch} (CIFAR-10) & 2 \\        \arrayrulecolor{black}\bottomrule
    \end{tabular}
   }
    \label{tab:gpu_days}
\end{table}
\section{Conclusions}
This paper introduced a new Cellular Encoding representation (XC) and a new ECNAS algorithm (XC-NAS), capable of evolving multi-path CNN architectures for image and text classification tasks. The XC representation is configurable and includes operations that can automatically modify a CNN architecture's width and depth during construction. This ability results in performant multi-path CNN architectures of varying complexity and capacity. XC-NAS implements a surrogate approach where only a subset of the training dataset is used during the evolutionary process, thereby evolving low-resolution models to save on compute costs and training time, after which the best-evolved model is retrained using the entire training set to produce a higher-resolution model for conducting inference. This approach has significantly reduced the required evolutionary run time. Experiments across five datasets show that the XC encoding is capable of representing a multitude of performant multi-path CNN architectures and that XC-NAS generalises well across both the image and text domains, demonstrating promising performance by outperforming several state-of-the-art approaches both in terms of classification performance and GPU days required to evolve the best architectures. Currently, XC-NAS can use a predefined micro-architecture of any configuration; however, future work will focus on integrating the ability to automatically evolve micro-architectures in conjunction with evolving the macro-architecture of a CNN architecture.

\bibliographystyle{splncs04}
\bibliography{references}

\begin{thebibliography}{10}
\providecommand{\url}[1]{\texttt{#1}}
\providecommand{\urlprefix}{URL }
\providecommand{\doi}[1]{https://doi.org/#1}

\bibitem{Broni-Bediako2020EvolutionaryEncoding}
Broni-Bediako, C., Murata, Y., Mormille, L.H.B., Atsumi, M.: {Evolutionary NAS
  with Gene Expression Programming of Cellular Encoding}. In: 2020 IEEE
  Symposium Series on Computational Intelligence (SSCI). pp. 2670--2676 (2020)

\bibitem{Clanuwat2018DeepLiterature}
Clanuwat, T., Bober-Irizar, M., Kitamoto, A., Lamb, A., Yamamoto, K., Ha, D.:
  {Deep Learning for Classical Japanese Literature} (2018)

\bibitem{Conneau2017VeryClassification}
Conneau, A., Schwenk, H., Cun, Y.L., Barrault, L.: {Very deep convolutional
  networks for text classification}. In: 15th Conference of the European
  Chapter of the Association for Computational Linguistics, EACL 2017 -
  Proceedings of Conference. vol.~2, pp. 1107--1116. Association for
  Computational Linguistics (ACL) (2017)

\bibitem{FernandesJunior2019ParticleClassification}
Fernandes~Junior, F.E., Yen, G.G.: {Particle swarm optimization of deep neural
  networks architectures for image classification}. Swarm and Evolutionary
  Computation pp. 62--74 (2019)

\bibitem{Gruau1994}
Gruau, F., Gruau, F., I, L.C.B.l., De~Doctorat, O.A.D., Demongeot, M.J.,
  Cosnard, E.M.M., Mazoyer, M.J., Peretto, M.P., Whitley, M.D.: {Neural Network
  Synthesis Using Cellular Encoding And The Genetic Algorithm.}  (1994)

\bibitem{He2016DeepRecognition}
He, K., Zhang, X., Ren, S., Sun, J.: {Deep residual learning for image
  recognition}. Proceedings of the IEEE Computer Society Conference on Computer
  Vision and Pattern Recognition pp. 770--778 (12 2016)

\bibitem{Holland1992GeneticAlgorithms}
Holland, J.H.: {Genetic Algorithms}. Scientific American  \textbf{267}(1),
  66--73 (1992)

\bibitem{Huang2021ADesign}
Huang, J., Xue, B., Sun, Y., Zhang, M.: {A Flexible Variable-length Particle
  Swarm Optimization Approach to Convolutional Neural Network Architecture
  Design}. 2021 IEEE Congress on Evolutionary Computation, CEC 2021 -
  Proceedings pp. 934--941 (2021)

\bibitem{KennedyParticleOptimization}
Kennedy, J., Eberhart, R.: {Particle swarm optimization}. Proceedings of
  ICNN'95 - International Conference on Neural Networks  \textbf{4},
  1942--1948

\bibitem{Koza1994GeneticSelection}
Koza, J.R.: {Genetic programming as a means for programming computers by
  natural selection}. Statistics and Computing 1994 4:2  \textbf{4}(2),
  87--112 (6 1994)

\bibitem{Krizhevsky2009LearningImages}
Krizhevsky, A., Hinton, G.: {Learning multiple layers of features from tiny
  images}. Tech. Rep.~0 (2009)

\bibitem{Londt2021EvolvingProgramming}
Londt, T., Gao, X., Andreae, P.: {Evolving Character-Level DenseNet
  Architectures Using Genetic Programming}. Lecture Notes in Computer Science
  pp. 665--680 (2021)

\bibitem{Poli2008AProgramming}
Poli, R., Langdon, W.B., McPhee, N.F.: {A Field Guide to Genetic Programming}.
  Lulu Enterprises, UK Ltd (2008)

\bibitem{Ren2019Eigen:Scratch}
Ren, J., Li, Z., Yang, J., Xu, N., Yang, T., Foran, D.J.: {Eigen:
  Ecologically-inspired genetic approach for neural network structure searching
  from scratch}. Proceedings of the IEEE Computer Society Conference on
  Computer Vision and Pattern Recognition pp. 9051--9060 (6 2019)

\bibitem{Suganuma2017AArchitectures}
Suganuma, M., Shirakawa, S., Nagao, T.: {A Genetic Programming Approach to
  Designing Convolutional Neural Network Architectures}. In: Proceedings of the
  Genetic and Evolutionary Computation Conference. pp. 497--504. GECCO '17
  (2017)

\bibitem{Sun2020EvolvingClassification}
Sun, Y., Xue, B., Zhang, M., Yen, G.G.: {Evolving Deep Convolutional Neural
  Networks for Image Classification}. IEEE Transactions on Evolutionary
  Computation (2),  394--407 (4 2020)

\bibitem{Szegedy2015GoingConvolutions}
Szegedy, C., Liu, W., Jia, Y., Sermanet, P., Reed, S., Anguelov, D., Erhan, D.,
  Vanhoucke, V., Rabinovich, A.: {Going deeper with convolutions}. Proceedings
  of the IEEE Computer Society Conference on Computer Vision and Pattern
  Recognition pp.~1--9 (10 2015)

\bibitem{Xiao2017Fashion-MNIST:Algorithms}
Xiao, H., Rasul, K., Vollgraf, R.: {Fashion-MNIST: a Novel Image Dataset for
  Benchmarking Machine Learning Algorithms}. CoRR  (2017)

\bibitem{Xie2017GeneticCNN}
Xie, L., Yuille, A.: {Genetic CNN}. In: 2017 IEEE International Conference on
  Computer Vision (ICCV). pp. 1388--1397 (2017)

\bibitem{Zhan2022EvolutionarySurvey}
Zhan, Z.H., Li, J.Y., Zhang, J.: {Evolutionary deep learning: A survey}.
  Neurocomputing pp. 42--58 (2022)

\bibitem{10.5555/2969239.2969312}
Zhang, X., Zhao, J., LeCun, Y.: {Character-Level Convolutional Networks for
  Text Classification}. In: Proceedings of the 28th International Conference on
  Neural Information Processing Systems - Volume 1. p. 649–657. NIPS'15, MIT
  Press, Cambridge, MA, USA (2015)

\bibitem{Zhou2021ASearch}
Zhou, X., Qin, A.K., Sun, Y., Tan, K.C.: {A Survey of Advances in Evolutionary
  Neural Architecture Search}. 2021 IEEE Congress on Evolutionary Computation,
  CEC 2021 - Proceedings pp. 950--957 (2021)

\end{thebibliography}

\end{document}